\documentclass[10pt,twocolumn,letterpaper]{article}

\usepackage{cvpr}              

\usepackage{graphicx}
\usepackage{amsmath}
\usepackage{amssymb}
\usepackage{booktabs}

\usepackage{amsmath,amsfonts}
\usepackage{algorithmic}
\usepackage{algorithm}
\usepackage{array}
\usepackage{textcomp}
\usepackage{stfloats}
\usepackage{url}
\usepackage{verbatim}
\usepackage{graphicx}
\usepackage{cite}

\newcommand{\widthscalefive}{0.3}
\usepackage{booktabs}
\usepackage{multirow}
\usepackage{color}
\usepackage{bbding}
\usepackage{adjustbox}
\usepackage{makecell}

\usepackage{color,xcolor,colortbl}
\definecolor{lightgray}{gray}{0.95}
\definecolor{color3}{gray}{0.95}
\definecolor{rouse}{rgb}{0.981,0.961,0.941}
\usepackage[accsupp]{axessibility}

\usepackage[pagebackref,breaklinks,colorlinks]{hyperref}

\usepackage[capitalize]{cleveref}
\crefname{section}{Sec.}{Secs.}
\Crefname{section}{Section}{Sections}
\Crefname{table}{Table}{Tables}
\crefname{table}{Tab.}{Tabs.}

\begin{document}


\title{Structured Sparsity Learning for Efficient Video Super-Resolution}
\author{%
	Bin Xia $^{1}$, Jingwen He $^{2}$, Yulun Zhang $^{3}$, Yitong Wang $^{4}$, \\ Yapeng Tian $^5$, Wenming Yang $^1$\thanks{Corresponding Author},  and Luc Van Gool $^3$ \\
	$^{1}$ Tsinghua University, $^2$  Shanghai AI Laboratory, $^3$ ETH Z\"{u}rich, \\$^4$  ByteDance Inc,  $^5$ University of Texas at Dallas
}
\maketitle

\vspace{-3mm}
\begin{abstract}
\vspace{-5mm}
The high computational costs of video super-resolution (VSR) models hinder their deployment on resource-limited devices, \eg, smartphones and drones. Existing VSR models contain considerable redundant filters, which drag down the inference efficiency. To prune these unimportant filters, we develop a structured pruning scheme called Structured Sparsity Learning (SSL) according to the properties of VSR. In SSL, we design pruning schemes for several key components in VSR models, including residual blocks, recurrent networks, and upsampling networks. Specifically, we develop a Residual Sparsity Connection (RSC) scheme for residual blocks of recurrent networks to liberate pruning restrictions and preserve the restoration information. For upsampling networks, we design a pixel-shuffle pruning scheme to guarantee the accuracy of feature channel-space conversion. In addition, we observe that pruning error would be amplified as the hidden states propagate along with recurrent networks. To alleviate the issue, we design Temporal Finetuning (TF). Extensive experiments show that SSL can significantly outperform recent methods quantitatively and qualitatively. The code is available at \url{https://github.com/Zj-BinXia/SSL}.  
\end{abstract}

\vspace{-6mm}
\section{Introduction}
\vspace{-3mm}
Video super-resolution (VSR) aims to generate a high-resolution (HR) video from its corresponding low-resolution (LR) observation by filling in missing details. With the popularity of intelligent edge devices such as smartphones and small drones, performing VSR on these devices is in high demand. Although a variety of VSR networks~\cite{EDVR,PFNL,3DCNN1,3DCNN2,li2019fast} can achieve great performance, these models are usually difficult to be deployed on edge devices with limited computation and memory resources. 

To alleviate this issue, we explore a new direction for effective and efficient VSR. To reduce the redundancy of Conv kernels~\cite{redundant1,redundant2,redundant3,redundant4} obtaining a more efficient VSR network, we develop a neural network pruning scheme for the VSR task for the first time. Since structured pruning~\cite{PL1,L2,SP1,ASSL} (focusing on filter pruning) can achieve an actual acceleration~\cite{L2,accelerate} superior to unstructured pruning~\cite{USP1,USP2} (focusing on weight-element pruning), we adopt structured pruning principle to develop our VSR pruning scheme. Given a powerful VSR network, our pruning scheme can find submodels under presetting pruning rate without significantly compromising performance.

\begin{figure*}[t]
	\centering
	\resizebox{0.88\linewidth}{!}{
	\includegraphics[height=6.8cm]{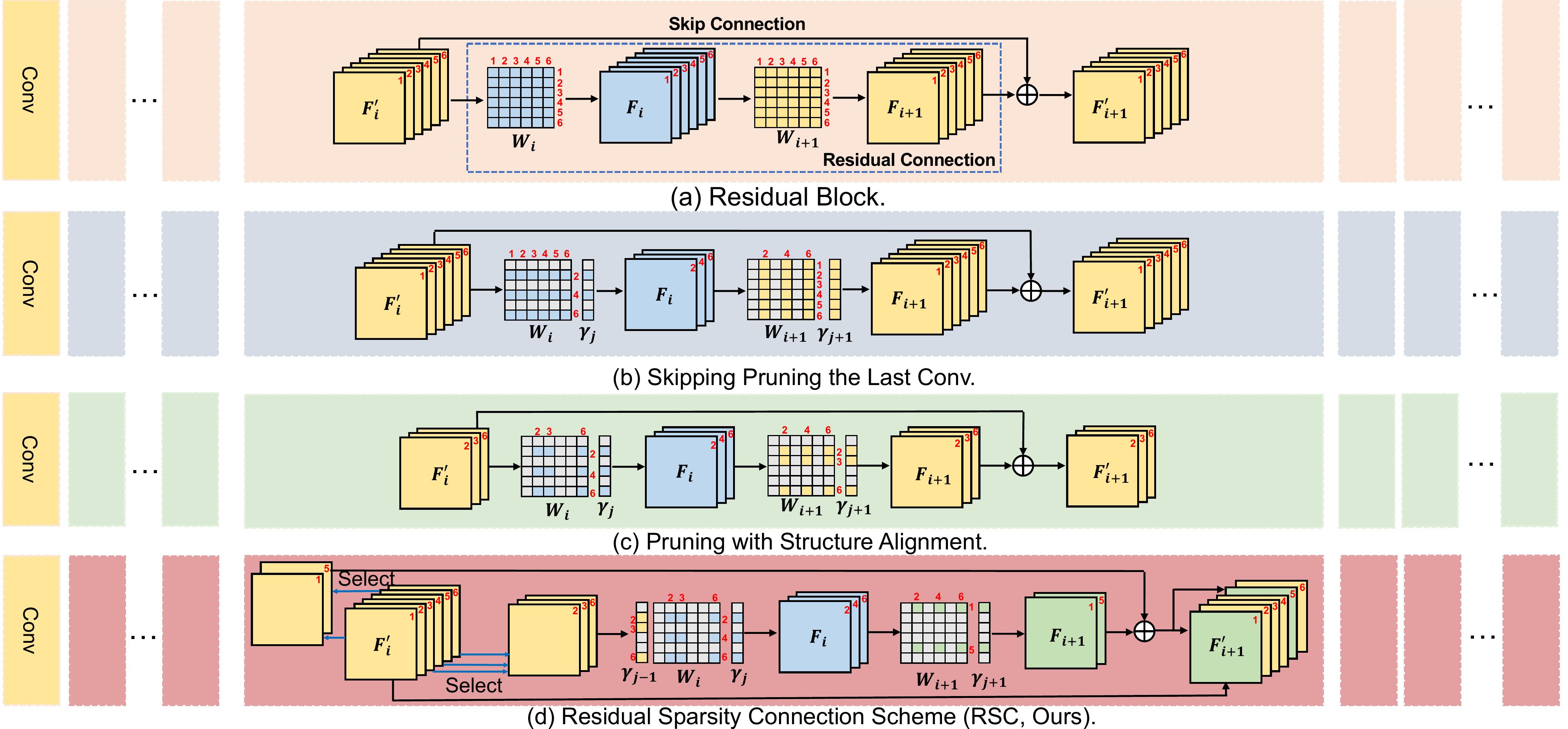}
	}
	\vspace{-3mm}
	\caption{Illustration of different
		schemes for pruning residual blocks of recurrent networks. \textbf{(a)} Structure of the residual block in the VSR network. \textbf{(b)} The residual block pruning schemes~\cite{PL1,PL2,PL3} do not prune the last Conv. \textbf{(c)} ASSL~\cite{ASSL} and SRPN~\cite{SRPN} prunes the same indices on skip and residual connections to keep channel alignment, which abandons some channels of input and output feature maps.  \textbf{(d)} RSC preserves all channels of input and output feature maps, which does not need to align the pruned indices on the first and last Convs in recurrent networks, can fully use restoration information, and can prune the first and last Convs of residual blocks without restrictions. }
		\vspace{-4mm}
	\label{fig:head}
\end{figure*}

 Structured pruning is a general concept, and designing a concrete pruning scheme for VSR networks is challenging. \textbf{(1)} Recurrent networks are widely used in VSR models to extract temporal features, consisting of residual blocks (\eg, BasicVSR~\cite{BasicVSR} has 60 residual blocks). However, it is hard to prune the residual blocks because the skip and residual connections ought to share the same indices~\cite{PL1} (Fig.~\ref{fig:head}~(a)).  As shown in Fig.~\ref{fig:head}~(b), quite a few structured pruning schemes~\cite{PL1,SLS} do not prune the last Conv layer of the residual blocks, which restricts the pruning space. Recently, as shown in Fig.~\ref{fig:head}~(c), ASSL~\cite{ASSL} and SRPN~\cite{SRPN} introduce regularization and prune the same indices on skip and residual connections to keep channel alignment (local pruning scheme,\ie, each layer pruning the same ratio of filters). However, ASSL and SRPN still cannot achieve the potential of pruning residual blocks on recurrent networks. The recurrent networks take the previous output as later input (Fig.~\ref{fig:process}~(a)). This requires the pruned indices of the first and last Convs in recurrent networks to be the same. But ASSL and SRPN cannot guarantee filter indices are aligned. Besides, many SR methods~\cite{ESRGAN,RDN} have shown that the information contained in front Conv layers can help the restoration feature extraction of later Conv layers. Thus, we design a Residual Sparsity Connection (RSC) for VSR recurrent networks, which preserves all channels of the input and output feature maps and selects the important channels for operation (Fig.~\ref{fig:head}~(d)). Compared with other pruning schemes~\cite{ASSL,SRPN}, RSC does not require the pruned indices of the first and last Convs of recurrent networks to be the same, can preserve the information contained in all layers, and liberates the pruning space of the last Conv of the residual blocks without adding extra calculations. Notably, RSC can prune residual blocks globally (\ie, the filters in various layers are compared together to remove unimportant ones).

 \textbf{(2)} We observe that the upsampling network accounts for $22\%$ of the total calculations in BasicVSR~\cite{BasicVSR}, which is necessary to be pruned to reduce redundancy. Since the pixel-shuffle~\cite{pixel-shuffle} operation in VSR networks converts the channels to space, pruning the pixel-shuffle without any restrictions would cause the channel-space conversion to fail. Thus, we specially design a pixel-shuffle pruning scheme by taking four consecutive filters as the pruning unit for 2$\times$ pixel-shuffle. \textbf{(3)} Furthermore, we observe that the error of pruning VSR networks would accumulate with propagation steps increasing along with recurrent networks, which limits the efficiency and performance of pruning. Thus, we further introduce Temporal Finetuning (TF) to constrain the pruning error accumulation in recurrent networks. Overall, our main contributions are threefold:
\begin{itemize}
    \vspace{-1mm}
	\item Our work is necessary and timely. There is an urgent need to compress VSR models for deployment. To the best of our knowledge, we are one of the first to design a structured pruning scheme for VSR. 
	\vspace{-1mm}
	\item We propose an integral VSR pruning scheme called Structured Sparsity Learning (SSL) for various components of VSR models, such as residual blocks, recurrent networks, and pixel-shuffle operations.  
	\vspace{-1mm}
	\item  We employ SSL to train VSR models, which surpass recent pruning schemes and lightweight VSR models.
\end{itemize}

\vspace{-2mm}
\section{Related Work}
\subsection{Video Super-Resolution}

VSR models can exploit additional information from neighboring LR frames for restoration~\cite{VSR2,VSR3,VSR4,VSR5,VSR6,VSR7,VSR8,basicvsr++,MANA,ETDM}. Earlier VSR methods~\cite{VSR9,VSR10,VSR6} estimate the optical flow between LR frames and perform spatial warping for alignment. Later methods resort to a more sophisticated approach of implicit alignment. Instead of image-level motion alignment, TDAN~\cite{TDAN} and EDVR~\cite{EDVR} work at the feature level. TDAN~\cite{TDAN} first adopted deformable Conv~\cite{deformable} in VSR to align the features of different frames. EDVR~\cite{EDVR} extended TDAN by introducing coarse-to-fine deformable alignment and a new spatial-temporal attention fusion module. RSDN~\cite{VSR2} adopted a recurrent detail-structural block and a hidden state adaptation module to reduce the effect of appearance changes and error accumulation. Recently, BasicVSR~\cite{BasicVSR} found that bidirectional propagation coupled with a simple optical flow-based feature alignment can further improve performance. Similarly, Yi~\etal~\cite{Omniscient} used the bidirectional propagation framework to exploit LR frames and estimated hidden states from the past, present, and future. To compress the VSR model, Xiao~\etal designed a space-time knowledge distillation scheme~\cite{distill-VSR}. However, these VSR methods require high computational costs impeding their application on resource-limited devices. Different from previous methods, we focus on designing SSL to compress VSR models by pruning redundant filters.

\vspace{-1mm}
\subsection{Network Pruning}
\vspace{-1mm}
Network pruning~\cite{redundant1,redundant2,redundant3,redundant4,SLS} is widely used to remove a set of redundant parameters for network acceleration. Pruning methods can be divided into two branches, structured pruning~\cite{PL1,L2,SP1,gao2018dynamic} and unstructured pruning~\cite{USP1,USP2}.  Structured pruning methods prune the network at the level of filters, channels, and even layers, which can obtain regular sparsity after pruning. This is beneficial for acceleration. In contrast, unstructured methods focus on pruning weights, leading up to much irregular sparsity. This is beneficial for compression but tends not to yield an actual acceleration~\cite{L2,accelerate}. Specifically, Li~\etal~\cite{PL1} applied the L$_{1}$-norm to measure the importance of different filters and then removed the less important ones.
Afterward, Liu~\etal~\cite{slimming} added a sparsity-inducing penalty term on scaling factors of the batch normalization layers to enforce the channels with lower scaling factors to be the less informative ones. Recently, ASSL~\cite{ASSL} and SRPN~\cite{SRPN} utilized aligned structured sparsity learning for structured pruning of residual blocks. In addition, Luo~\etal~\cite{luo2020neural} developed a residual block pruning scheme for image classification using the Convs on skip connections. However, the residual blocks of VSR networks do not have such Convs. Lin~\etal~\cite{lin2017runtime} conducted runtime neural network pruning according to the input image. Besides, Wang~\etal~\cite{SMSR} designed an unstructured pruning scheme for single image SR tasks by using sparse Conv to skip redundant computations. Since we cannot directly apply a general structured pruning method for VSR, we explore the properties of VSR networks and develop a VSR pruning scheme in this paper.  

\vspace{-2mm}
\section{Methodology}
\label{sec:method}
\vspace{-1mm}

\subsection{Overview}
\begin{figure*}[t]
	\centering
	\resizebox{1\linewidth}{!}{
	\includegraphics[height=4.8cm]{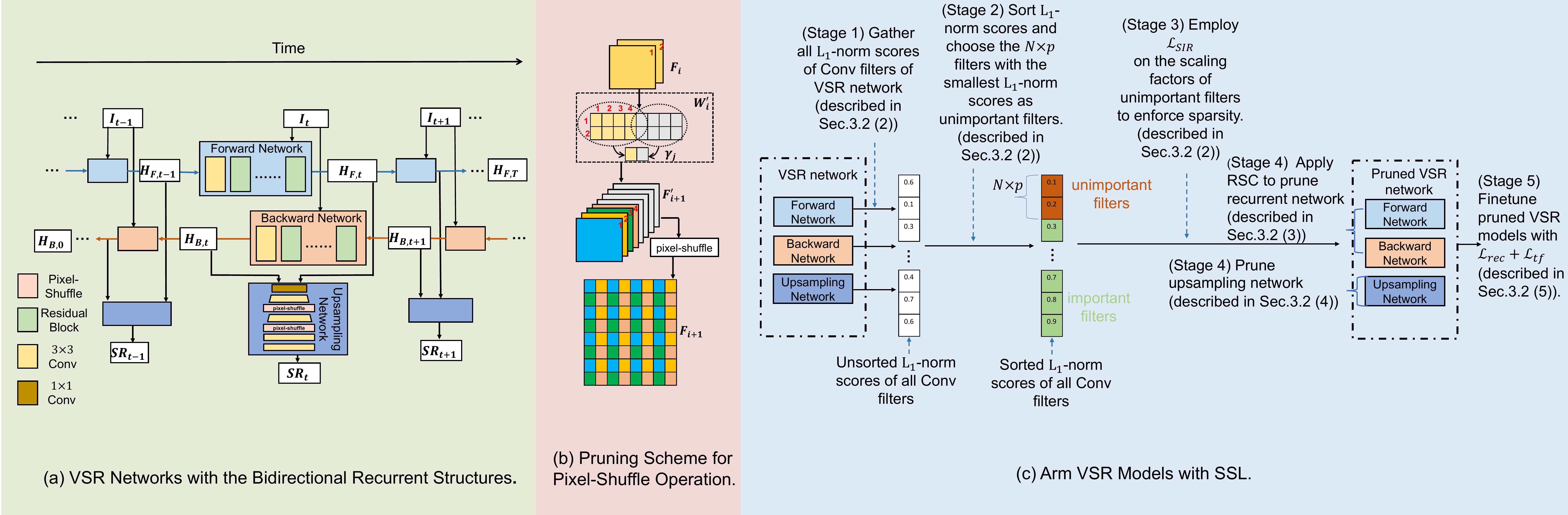}
	}
	\vspace{-6mm}
	\caption{\textbf{(a)} The basic architecture of the VSR methods with the bidirectional recurrent network. The forward and backward networks both consist of numerous residual blocks. The upsampling networks contain multiple pixel-shuffle operations and Convs. \textbf{(b)} The pruning scheme for the pixel-shuffle. For the $2\times$ upsampling pixel-shuffle~\cite{pixel-shuffle} operation, we take four channels with consecutive indices as the pruning unit to guarantee the accuracy of channel-space conversion after pruning. \textbf{(c)} The application of SSL on VSR models.}
	\label{fig:process}
    \vspace{-6mm}
\end{figure*}

Figure~\ref{fig:process}~(a) shows VSR networks based on the bidirectional recurrent structures, such as BasicVSR~\cite{BasicVSR}. Given a LR frame $I_{t}$, the forward network concatenates $I_{t}$ and the previous hidden state $H_{F,t-1}$ to extract features from $I_{t}$ and aggregate the reference information from $H_{F,t-1}$. Similarly, the backward network extracts features from $I_{t}$ and aggregates the reference information from the future hidden state $H_{B,t+1}$. Note that both the forward and backward networks consist of numerous residual blocks. Then, the features generated by forward and backward networks are fed into the upsampling network, which consists of multiple pixel-shuffle operations and Convs, to obtain the recovered frame $SR_{t}$. However, SOTA VSR networks~\cite{BasicVSR,basicvsr++,Omniscient} need massive computation and memory resources, limiting their deployment on resource-limited devices.  

To pursue more efficient VSR networks, we specially design a VSR structured pruning scheme called Structured Sparsity Learning (SSL), according to the properties of VSR networks. Specifically, SSL has three stages, including pretraining, pruning, and finetuning. In the pretraining stage, we train a powerful VSR network. Since current VSR networks do not use BatchNorm~\cite{BN}, we introduce a scaling factor in pretrained VSR models to tune the sparsity of each channel and filter. In the pruning stage, we select the unimportant filters according to the pruning criterion and apply sparsity-inducing regularization on corresponding scaling factors. In addition,  we propose a Residual Sparsity Connection (RSC) scheme to liberate the restrictions on pruning residual blocks of recurrent networks and preserve all restoration information contained in channels of feature maps for better performance. Moreover, for the upsampling networks, we specially develop a pruning scheme for the pixel-shuffle operation to guarantee the accuracy of channel-space conversion after pruning. Besides, we observe that the error of the hidden state would be amplified with the propagation in recurrent networks after pruning. Thus, in the finetuning stage, we design Temporal Finetuning (TF) to alleviate the error accumulation.

\subsection{Structured Sparsity Learning}
\vspace{-1mm}
\label{sec:res}
Structured Sparsity Learning (SSL) is a structured pruning scheme specially designed for VSR. It can reduce the redundancy of neural filters and obtain more efficient VSR submodels.  Next, we will explain our SSL in detail.

\textbf{(1) Scaling Factor.}    
Structured pruning aims to remove Conv filters based on a designed importance criterion. In the classification task, quite a few works use scale parameters of BatchNorm~\cite{BN} to control the throughput of each filter. Zero scale parameters make the value of corresponding channels vanish. As a result, they contribute nothing to the subsequent Convs and can be removed. By regularizing the scale parameter, we can assess and tune the importance of each filter. However, the BatchNorm is not useful for SR tasks~\cite{EDSR}, and SOTA VSR networks~\cite{BasicVSR,basicvsr++,Omniscient} do not utilize it. Therefore, it is infeasible to apply the existing pruning schemes directly. In our pruning scheme, as shown in Fig.~\ref{fig:head} (d) and Fig.~\ref{fig:process} (b), we multiply the scaling factors $\boldsymbol{\gamma}$ before or after Convs. Then, we perform regularization on scaling factors to enforce sparsity for pruning.

\textbf{(2) Pruning Criterion and Regularization Form.}
To remove the redundant filters, we need to select unimportant scaling factors $\boldsymbol{\gamma}$ to induce sparsity. In previous works, ASSL~\cite{ASSL} and SRPN~\cite{SRPN} had to adopt a local pruning scheme (namely, scaling factors are only compared within the same layer, and each layer has the same pruning ratio) to guarantee that skip and residual connections keep the same number of filters and indices for the adding operation. Given that the importance of the Conv filters in various layers is different and our RSC does not have restrictions as ASSL, we can adopt the global pruning scheme (\ie, scaling factors of different layers are compared together). 

For the pruning criterion, we adopt the simple and practical L$_{1}$-norm. Specifically, given Conv kernel $\boldsymbol{W_{i}}\in \mathbb{R}^{ C_{out} \times C_{in} \times K_{h} \times K_{w}}$ in the $i$-th layer, we calculate the absolute weight sum of $k$-th Conv filter $\boldsymbol{W_{i}}[k,...] \in \mathbb{R}^{ C_{in} \times K_{h} \times K_{w}}$ with $s_{i,k}=\sum\left|\boldsymbol{W_{i}}[k,...]\right|$. In particular, for our RSC in Fig.~\ref{fig:head} (d), we require to additionally prune the input channels for the first Conv, and calculate its L$_{1}$-norm score with $s_{i,k}^{\prime}= \sum\left|\boldsymbol{W_{i}}[:,k,...]\right|$, where $\boldsymbol{W_{i}}[:,k,...] \in \mathbb{R}^{ C_{out} \times K_{h} \times K_{w}}$. Moreover, for the Conv before the pixel-shuffle operation, we take four consecutive filters as a pruning unit and calculate the score with $s_{i,k}= \sum\left|\boldsymbol{W_{i}}[4k:4(k+1),...]\right|$. Then, given the pruning ratio $p$ and the total number of filters or channels $N$, we sort all their L$_{1}$-norm scores $s$ together and choose the $N\times p$ filters with the smallest L$_{1}$-norm values as unimportant filters or channels, denoted as set $S$ (Fig.~\ref{fig:process} (c)).  

After identifying the unimportant filters and channels set $S$, we apply sparsity-inducing regularization (SIR) to the corresponding scaling factors, denoted as set $S_{sf}$. It is notable that we do not enforce sparsity-inducing regularization to the important filters and channels since they will remain in the network. Specifically, we employ $L_{2}$ regularization on the scaling factors to enforce sparsity (Fig.~\ref{fig:process} (c)):
\begin{equation}
\vspace{-1mm}
\mathcal{L}_{SIR}=\alpha_{\gamma} \sum_{\gamma \in S_{sf}} \gamma^{2},
\label{eq:lsir}
\vspace{-2mm}
\end{equation}
where $\gamma$ is the scalar selected from $\boldsymbol{\gamma} \in \mathbb{R}^{C }$ corresponding to unimportant filters or channels; $\alpha_{\gamma}$ is a scalar. We increment $\alpha_{\gamma}$ by a presetting constant $\Delta$ every $T_{1}$ iterations. When $\alpha_{\gamma}$ reaches the pre-defined upper limit $\tau$, we keep $\alpha_{\gamma}$ constant and continue training $T_{2}$ iterations.

\textbf{(3) Pruning Scheme for Residual Blocks of Recurrent Networks.}  VSR recurrent networks consist of residual blocks. Residual blocks are difficult to prune because the addition operations require the pruned filter indices between the skip and residual connections to be the same. As shown in Fig.~\ref{fig:head}~(b), quite a few pruning schemes~\cite{PL1,slimming,SLS} simply skipped the pruning of the last Conv in residual blocks, which restricted the pruning space. Recently, as shown in Fig.~\ref{fig:head}~(c), ASSL~\cite{ASSL} and SRPN~\cite{SRPN} pruned the last Conv in the residual block by using regularization. However, recurrent networks take the previous output as later input. Thus, pruning recurrent networks require the pruned indices of the first and last Convs to be the same. ASSL and SRPN cannot guarantee the indices of important filters in the first and last Convs of recurrent networks are aligned. Besides, ASSL and SRPN have to remove some channels in output feature maps and adopt a local pruning scheme, which limits pruning space and restoration information utilization (RDN~\cite{RDN} and ESRGAN~\cite{ESRGAN} have shown restoration information of front layers can guide feature extraction of later layers). To break the restriction of pruning and fully use restoration information contained in channels of front layers, we propose the RSC to prune residual blocks in recurrent networks (Fig.~\ref{fig:head} (d)). As we can see, RSC preserves all channels of input $\boldsymbol{F_{i}^{\prime}}$ and output $\boldsymbol{F_{i+1}^{\prime}}$ in the residual blocks. For the first Conv, we select the important channels (the indices not in $S$) to participate in the Conv operation, which can be expressed as Eq.~\ref{Eq:res1}. After the last Conv, we obtained $\boldsymbol{F_{i+1}}$ and add $\boldsymbol{F_{i+1}}$ to $\boldsymbol{F_{i}^{\prime}}$ on the corresponding channel indices to obtain $\boldsymbol{F_{i+1}^{\prime}}$ (Eq.~\ref{Eq:res2} and Eq.~\ref{Eq:res3}). Furthermore, we do not prune the 1$\times$1 Conv of upsampling networks (Fig.~\ref{fig:process}~(a)) to aggregate all preserved restoration information in $H_{F,t},H_{B,t}\in \mathbb{R}^{C\times H \times W}$. 
\begin{align}
\boldsymbol{F_{i}} =  &\boldsymbol{F_{i}^{\prime}} \otimes (\boldsymbol{\gamma_{j-1}} \boldsymbol{W_{i}} \boldsymbol{\gamma_{j}}) ,
\label{Eq:res1} \\
\boldsymbol{F_{i+1}} &= \boldsymbol{F_{i}} \otimes (\boldsymbol{W_{i+1}} \boldsymbol{\gamma_{j+1}}),
\label{Eq:res2} \\
\boldsymbol{F_{i+1}^{\prime}} &= \boldsymbol{F_{i+1}} + \boldsymbol{F_{i}^{\prime}},
\label{Eq:res3} 
\end{align} 
where $\otimes$ indicates Conv. $\boldsymbol{F_{i}^{\prime}}, \boldsymbol{F_{i+1}^{\prime}} \in \mathbb{R}^{C\times H \times W}$ are the input and output feature maps of the residual block, respectively. $\boldsymbol{F_{i}}, \boldsymbol{F_{i+1}} \in \mathbb{R}^{C_{p}\times H \times W}$ are intermediate feature maps. $\boldsymbol{W_{i}},\boldsymbol{W_{i+1}} \in \mathbb{R}^{C_{out}\times C_{in} \times K_{h} \times K_{w}}$ are weights of Conv kernels. $\boldsymbol{\gamma_{j-1}} \in \mathbb{R}^{ C_{in}}$ and $\boldsymbol{\gamma_{j}}, \boldsymbol{\gamma_{j+1}} \in \mathbb{R}^{C_{out} }$ are scaling factors to apply sparsity-inducing regularization.  $\boldsymbol{F_{i+1}^{\prime}}$ prunes some channels, and $\boldsymbol{F_{i}}$ keeps all channels. In Eq.~\ref{Eq:res3}, $\boldsymbol{F_{i}^{\prime}}$ adds $\boldsymbol{F_{i+1}}$ on corresponding kept channels. It is notable that our RSC does not introduce extra parameters and computational cost.

\textbf{(4) Pruning Scheme for Pixel-Shuffle.} The upsampling network of the VSR network uses Conv to increase channels of feature maps and adopts the pixel-shuffle~\cite{pixel-shuffle} operation to convert the channels to space realizing upsampling. As shown in Fig.~\ref{fig:process} (b), given the input feature map $F_{i} \in \mathbb{R}^{C \times H \times W}$, we expand its channels $4\times$ by a Conv with weight $W_{i}$ to obtain $F_{i+1}^{\prime} \in \mathbb{R}^{4C \times H \times W}$. Then, the pixel-shuffle operation takes four channels as a group to convert $F_{i+1}^{\prime}$ to $F_{i+1} \in \mathbb{R}^{C \times 2H \times 2W}$ realize $2\times$ upsampling. We observe that, if we prune the Conv before pixel-shuffle without any restriction, the pruned feature maps will be spatially disordered after passing the pixel-shuffle operation and lead to performance drop. To address the problem, we specially design a strong and simple pruning scheme for the pixel-shuffle operation. Given the input feature map, we take four filters as a pruning unit to evaluate their importance and then impose the scaling factor $\boldsymbol{\gamma_{j}}$ on filters to enforce sparsity (described in (2) pruning criterion and regularization form):   
\vspace{-2mm}
\begin{equation}
\boldsymbol{W_{i}^{\prime}}=  \boldsymbol{W_{i}}[4k:4(k+1),...] \boldsymbol{\gamma_{j}}[k],  k \in [0,C_{in}),
\vspace{-1mm}
\label{Eq:up1}
\end{equation}  
where $\boldsymbol{W_{i}} \in \mathbb{R}^{4C_{in} \times C_{in} \times K_{h} \times K_{w}}$ is the weights of Conv kernel. $\boldsymbol{\gamma_{j}}$ is the scaling factor. 

\textbf{(5) Temporal Finetuning.} 
We observe that the pruned VSR network generates a minor error in hidden state $H_{F}$ and $H_{B}$ (Fig.~\ref{fig:process} (a)), which will be amplified as the hidden state propagates along with recurrent networks. To solve the issue, we introduce Temporal Finetuning (TF) by enforcing the hidden states of pruned networks to align with the accurate  hidden states of unpruned networks:
\vspace{-1mm}
\begin{equation}
\mathcal{L}_{tf}=\left\|H_{F,T}-H_{F,T}^{\prime}\right\|+\left\|H_{B,0}-H_{B,0}^{\prime}\right\|,
\label{eq:ltf}
\vspace{-2mm}
\end{equation}  
where $T$ is the number of input frames, and $H_{F,T}$ and $H_{F,T}^{\prime}$ are the final hidden states after $T$ frames forward propagation in pruned and original VSR networks, separately. Similarly, $H_{B,0}$ and $H_{B,0}^{\prime}$ are the final hidden states after backward propagation in the pruned and original VSR networks.

To train the whole VSR network, we use the Charbonnier loss~\cite{BasicVSR,EDVR}, which can be formulated as:
\vspace{-1mm}
\begin{equation}
\mathcal{L}_{r e c}=\frac{1}{T}\sum_{t=1}^{T}\sqrt{\left\|S R_{t}-H R_{t}\right\|^{2}+\varepsilon^{2}},
\label{eq:lrec}
\vspace{-2mm}
\end{equation}
where $\varepsilon$ is set to $10^{-6}$. $SR_{t}$ and $HR_{t}$ are $t$-th reconstructed and HR frames, respectively. The overall loss function for pruned network finetuning is designed as:
\begin{equation}
\mathcal{L}_{all}=\mathcal{L}_{rec}+ \mathcal{L}_{tf}.
\label{eq:lall}
\end{equation}

\subsection{Arm VSR Models with SSL}
\vspace{-1mm}
Our SSL can be used for VSR networks. BasicVSR~\cite{BasicVSR} and BasicVSR++~\cite{basicvsr++} are two strong SOTA VSR methods. The dense deformable Conv of BasicVSR++ requires reading a large amount of irregular memory data, which is unsuitable for deployment on resource-limited devices without GPUs. Besides, the Second-Order Grid Propagation of BasicVSR++ will force a delay of two frames, which further impedes its usage in real-time devices.

In our study, we use BasicVSR for VSR pruning, which is more suitable for applications on edge devices. In addition, we further propose unidirectional BasicVSR (BasicVSR-uni), obtained by removing the backward network, for online inference. Since the SpyNet of BasicVSR is used for flow estimation, we do not apply our pruning scheme SSL on it. In the pruning stage, as shown in Fig.~\ref{fig:process} (c), we first add the scaling factor to the Conv and residual blocks as described in Sec.~\ref{sec:res}. Then we use the pruning criterion to select unimportant filters globally and apply sparsity-inducing regularization to the corresponding scaling factor. Afterward, we remove the unimportant Conv filters and finetune the pruned VSR network with $T_{3}$ iterations. We provide more details in Alg.~\textcolor{red}{1} of supplementary.         

\vspace{-2mm}
\section{Experiments}
\vspace{-1mm}
\subsection{Experimental Settings} 
\vspace{-1mm}
\label{sec:setting}
We adopt two widely used datasets for training: REDS~\cite{REDS} and Vimeo-90K~\cite{VSR6}. For REDS, we use REDS4 containing $4$ clips as our test set. Additionally, we adopt REDSval4 as our validation set, which contains $4$ clips selected from the REDS validation set. The remaining clips of REDS are used for training. In addition, we utilize Vid4~\cite{Vid4}, UDM10~\cite{VSR7}, and Vimeo-90K-T~\cite{VSR6} as test sets along with Vimeo-90K. We train and test models with 4$\times$ downsampling using two degradations Bicubic (BI) and Blur Downsampling (BD) as BasicVSR did. For BI, the MATLAB function ``imresize" is used for downsampling. For BD, we blur the HR images by a Gaussian filter with $\sigma=$1.6, followed by a subsampling every four pixels.

We pretrain the unidirectional BasicVSR (BasicVSR-uni) as done for BasicVSR. In sparsity-inducing regularization, the iterations $T_{1}$ and $T_{2}$ are set to $5$ and $3,375$ separately. The scalars $\Delta$ and $\tau$ are set to $10^{-4}$ and $0.1$, respectively. Note that we fix the parameters of the flow estimator in sparsity-inducing regularization. In the pruned VSR network finetuning, we set $T_{3}$ to $300,000$. We adopt the Adam optimizer~\cite{Adam} and Cosine Annealing scheme~\cite{cosine}. The initial learning rate of the flow estimator is $2.5\times10^{-5}$. The learning rate for all other modules is $2 \times 10^{-4}$. The patch size of input LR frames is $64\times64$.  Experiments are conducted
on a server with PyTorch 1.10 and V100 GPUs.

\begin{table*}[t]
\Huge
	\centering
		\caption{Quantitative comparison (average PSNR/SSIM). Pruning schemes applied on bidirectional and unidirectional BasicVSR (``bi'' and ``uni'') and marked in rouse and gray, respectively. $*$ means the space-time knowledge distillation scheme~\cite{distill-VSR}. We mark the best results among comparing pruning schemes in bold. The FLOPs and runtime are computed based on an LR size of $180\times320$.}
		\vspace{-2mm}
	\resizebox{0.98\linewidth}{!}{
	\begin{tabular}{cccc|ccc|ccc}
		\toprule[0.2em]
		&       &       &       & \multicolumn{3}{c|}{BI degradatioin} & \multicolumn{3}{c}{BD degradatioin} \\
		Methods & Params (M) & FLOPs (G) & Runtime (ms) & REDS4~\cite{REDS} & Vimeo-90K-T~\cite{VSR6} & Vid4~\cite{Vid4}  & UDM10~\cite{VSR7} & Vimeo-90K-T~\cite{VSR6} & Vid4~\cite{Vid4} \\
		 \midrule
		Bicubic & -     & -     & -     & 26.14/0.7292 & 31.32/0.8684 & 23.78/0.6347 & 28.47/0.8253 & 31.30/0.8687 & 21.80/0.5246 \\
		VESPCN~\cite{VSR9} & -     & -     & -     & -     & -     & 25.35/0.7557 & -     & -     & - \\
		SPMC~\cite{VSR10}  & -     & -     & -     & -     & -     & 25.88/0.7752 & -     & -     & - \\
		TOFlow~\cite{VSR6} & 1.4   & 274.9 & 1610  & 27.98/0.7990 & 33.08/0.9054 & 25.89/0.7651 & 36.26/0.9438 & 34.62/0.9212 & - \\
		DUF~\cite{DUF}   & 5.8   & 1645.8 & 974   & 28.63/0.8251 & -     & -     & 38.48/0.9605 & 36.87/0.9447 & 27.38/0.8329 \\
		RBPN~\cite{RBPN}  & 12.2  & 8516.0  & 1507  & 30.09/0.8590 & 37.07/0.9435 & 27.12/0.8180 & 38.66/0.9596 & 37.20/0.9458 & - \\
		EDVR-M~\cite{EDVR} & 3.3   & 304.2 & 118   &  30.53/0.8699   &    37.09/0.9446   &   27.10/0.8186 & 39.40/0.9663 & 37.33/0.9484 & 27.45/0.8406 \\
		PFNL~\cite{PFNL}  & 3.0     & 940.0   & 295   & 29.63/0.8502 & 36.14/0.9363 & 26.73/0.8029 & 38.74/0.9627 & -     & 27.16/0.8355 \\
		TGA~\cite{TGA}   & 5.8   & 694.1 & 236   & -     & -     & -     & -     & 37.59/0.9516 & 27.63/0.8423 \\
		RLSP~\cite{RLSP}  & 4.2   & 82.3  & 49    & -     & -     & -     & 38.48/0.9606 & 36.49/0.9403 & 27.48/0.8388 \\
		RSDN~\cite{VSR2}  & 6.2   & 355.7 & 94    & -     & -     & -     & 39.35/0.9653 & 37.23/0.9471 & 27.92/0.8505 \\
		RRN~\cite{RRN}   & 3.4   & 108.7 & 45    & -     & -     & -     & 38.96/0.9644 & -     & 27.69/0.8488 \\
		FastDVDnet$^{*}$~\cite{distill-VSR}  & 2.6   & 64.3 & - & - & 36.12/0.9348 & 26.14/0.8029 & - & - & - \\
		 \midrule
		\rowcolor{rouse}
		BasicVSR~\cite{BasicVSR} & 4.9   & 338.5 & 57    & 31.42/0.8909 & 37.18/0.9450 & 27.24/0.8251 & 39.96/0.9694 & 37.53/0.9498 & 27.96/0.8553 \\
		\rowcolor{rouse}
		BasicVSR-lite & 1.3   & 85.5 & 24    & 
		30.56/0.8738 & 36.57/0.9397	& 26.86/0.8125 &	38.98/0.9645 & 36.78/0.9431 &27.27/0.8327 \\
		\rowcolor{rouse}
		L$_{1}$-norm-bi~\cite{PL1} & 1.3   & 85.5 & 24    & 
		30.66/0.8766 & 36.69/0.9406 & 26.87/0.8121 &	39.04/0.9650 & 36.84/0.9437 & 27.29/0.8335 \\
		\rowcolor{rouse}
		ASSL-bi~\cite{ASSL} & 1.3   & 85.5 & 24    & 
		30.74/0.8770 & 36.75/0.9414 & 27.01/0.8176 &	39.15/0.9660 & 36.93/0.9450 &	27.40/0.8400 \\
		\rowcolor{rouse}
		SSL-bi (Ours) & 1.0   & 92.1 & 24    &
		\textbf{31.06/0.8933} & \textbf{36.82/0.9419} &	\textbf{27.15/0.8208} & \textbf{39.35/0.9665} & \textbf{37.06/0.9458} &	\textbf{27.56/0.8431} \\
		\midrule
		\rowcolor{lightgray}
		BasicVSR-uni~\cite{BasicVSR} & 2.6     & 218.1 & 39    & 30.56/0.8698 & 36.95/0.9429 & 27.01/0.8164 & 39.25/0.9645 & 37.25/0.9472 & 27.57/0.8424 \\
		\rowcolor{lightgray}
		BasicVSR-uni-lite & 0.7   & 62.4  & 18    & 29.95/0.8561 & 36.38/0.9372 & 26.68/0.8012 & 38.24/0.9586 & 36.38/0.9388 & 26.87/0.8157 \\
		\rowcolor{lightgray}
		L$_{1}$-norm-uni~\cite{PL1} & 0.7   & 62.4  & 18    & 29.97/0.8570 & 36.45/0.9381 &	26.70/0.8031 & 38.43/0.9601 & 36.53/0.9405 &	26.89/0.8187 \\
		\rowcolor{lightgray}
		ASSL-uni~\cite{ASSL} & 0.7   & 62.4  & 18    & 30.02/0.8589 & 36.49/0.9385 & 26.76/0.8051 & 38.48/0.9603 & 36.61/0.9416 & 27.02/0.8236 \\
		\rowcolor{lightgray}
		SSL-uni (Ours) & 0.5   & 63.9  & 18    & 
		\textbf{30.24/0.8633} & \textbf{36.56/0.9392} & \textbf{27.01/0.8148} &	\textbf{38.68/0.9615} & \textbf{36.77/0.9429} & \textbf{27.18/0.8296} \\
		\bottomrule[0.2em]
	\end{tabular}%
}
\vspace{-5mm}
	\label{tab:SR}%
\end{table*}%

\vspace{-1mm}
\subsection{Quantitative and Qualitative Comparisons}
\vspace{-1mm}
Since BasicVSR violates causality and cannot be evaluated online, we construct the unidirectional BasicVSR (BasicVSR-uni) by removing the backward network for online inference. We compare the proposed SSL with three other pruning schemes at pruning ratio $p=0.5$: L$_{1}$-norm pruning~\cite{PL1} (which simply removes filters with the smallest L$_{1}$-norms and is the most prevailing filter pruning method now), and ASSL~\cite{ASSL}. We apply these pruning schemes on BasicVSR, thus obtaining L$_{1}$-norm-bi, ASSL-bi, and SSL-bi separately. In addition, we use these pruning schemes on BasicVSR-uni, obtaining L$_{1}$-norm-uni, ASSL-uni, and SSL-uni. Moreover, we reduce the channels of BasicVSR and BasicVSR-uni to obtain lightweight VSR models BasicVSR-lite and BasicVSR-uni-lite, respectively.  Furthermore, we compare our pruned BasciVSR and BasicVSR-uni with other lightweight VSR networks, including TOFlow~\cite{VSR6}, EDVR-M~\cite{EDVR}, RLSP~\cite{RLSP}, RSDN~\cite{VSR2}, \etc. Since we only prune VSR networks, the parameters and FLOPs of the optical flow network, SPyNet~\cite{spynet} (Params 1.4M, Flops 19.6G) are not included.

The quantitative results measures performance (PSNR and SSIM), the number of parameters, runtime, and FLOPs on the different methods, which are shown in Tab.~\ref{tab:SR}. \textbf{(1)} Compared with competitive lightweight VSR networks, our SSL-bi obtains 0.53dB gain on REDS4 over EDVR-M. Note that, different from careful network designs like EDVR-M, we merely prune the BasicVSR, a simple backbone with 60 residual blocks, obtaining superior performance while only consuming 
less the FLOPs of EDVR-M. \textbf{(2)} Our SSL-bi surpasses the BasicVSR-lite by 0.5dB, and SSL-uni surpasses the BasicVSR-uni-lite by 0.29dB. This demonstrates the effectiveness of applying SSL for offline and online VSR network pruning. \textbf{(3)} We also adapt some SOTA pruning schemes, such as the L$_{1}$-norm and ASSL, to VSR networks for comparison. As a result, our SSL achieves superior performance on BasicVSR and BasicVSR-uni to other pruning schemes. This shows that SSL can make better use of the sparsity of the network and increases the efficiency of the learned network parameters. \textbf{(4)} Besides, comparing our pruning scheme and other VSR model compression method (space-time knowledge distillation scheme~\cite{distill-VSR}, FastDVDnet$^{*}$), our SSL-uni (0.5M parameters) surpasses the FastDVDnet$^{*}$ (2.6M parameters) by 0.87dB on Vid4, which further demonstrates the effectiveness of our structured pruning scheme. Moreover, with our SSL strategy, we do not have to train a teacher network as knowledge distillation~\cite{distill-VSR} methods did.

\begin{figure*}[tb]
	\LARGE
	\centering
	\resizebox{1\linewidth}{!}{
		\begin{tabular}{cc}
		\begin{adjustbox}{valign=t}
				\LARGE
				\begin{tabular}{c}
					\includegraphics[height=0.44\textwidth]{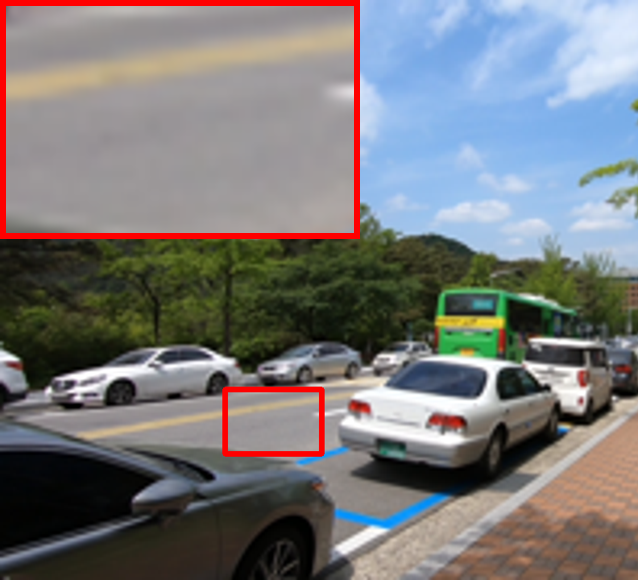} \\
					LR
				\end{tabular}
				
			\end{adjustbox}
			
			\begin{adjustbox}{valign=t}
				\LARGE
				\begin{tabular}{cccccc}
					\includegraphics[width=\widthscalefive \textwidth]{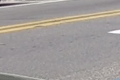}  &
					\includegraphics[width=\widthscalefive \textwidth]{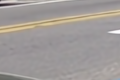}  &
					\includegraphics[width=\widthscalefive \textwidth]{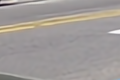}  &
					\includegraphics[width=\widthscalefive \textwidth]{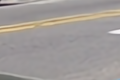}  &
					\includegraphics[width=\widthscalefive \textwidth]{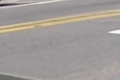}  &
					\includegraphics[width=\widthscalefive \textwidth]{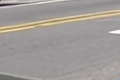} 
					\\
					HR  &
					\makecell{BasicVSR}  &
					\makecell{BasicVSR-lite}  &
					\makecell{L$_{1}$-norm-bi}  &
					\makecell{ASSL-bi}  &
					\makecell{SSL-bi (Ours)} 
					\\
					\includegraphics[width=\widthscalefive \textwidth]{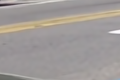}  &
					\includegraphics[width=\widthscalefive \textwidth]{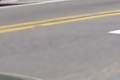}  &
					\includegraphics[width=\widthscalefive \textwidth]{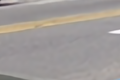}  &
					\includegraphics[width=\widthscalefive \textwidth]{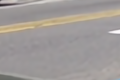}  &
					\includegraphics[width=\widthscalefive \textwidth]{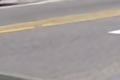}  &
					\includegraphics[width=\widthscalefive \textwidth]{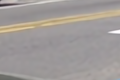} 
					\\ 
					EDVR-M  &
					\makecell{BasicVSR-uni}  &
					\makecell{BasicVSR-uni-lite}  &
					\makecell{L$_{1}$-norm-uni}  &
					\makecell{ASSL-uni}  &
					\makecell{SSL-uni (Ours)} 
				\end{tabular}
			\end{adjustbox}	
			\\
		\begin{adjustbox}{valign=t}
				\LARGE
				\begin{tabular}{c}
					\includegraphics[height=0.44\textwidth]{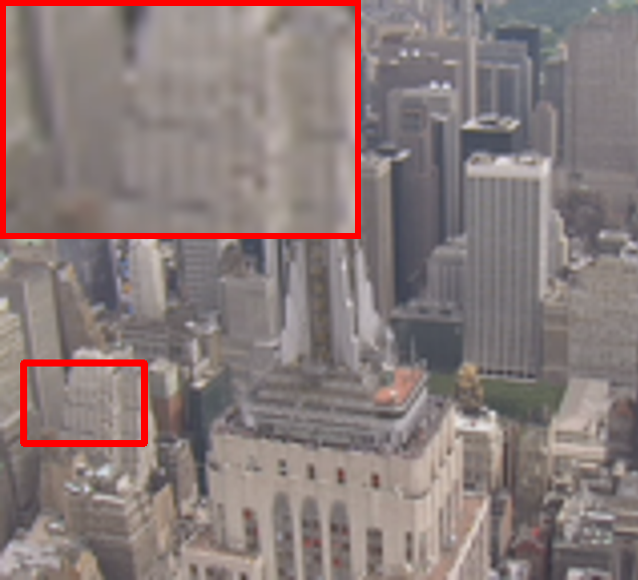} \\
					LR
				\end{tabular}
				
			\end{adjustbox}
			
			\begin{adjustbox}{valign=t}
				\LARGE
				\begin{tabular}{cccccc}
					\includegraphics[width=\widthscalefive \textwidth]{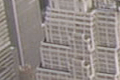}  &
					\includegraphics[width=\widthscalefive \textwidth]{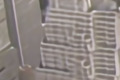}  &
					\includegraphics[width=\widthscalefive \textwidth]{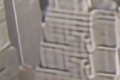}  &
					\includegraphics[width=\widthscalefive \textwidth]{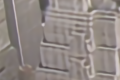}  &
					\includegraphics[width=\widthscalefive \textwidth]{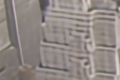}  &
					\includegraphics[width=\widthscalefive \textwidth]{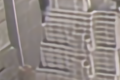} 
					\\
					HR  &
					\makecell{BasicVSR}  &
					\makecell{BasicVSR-lite}  &
					\makecell{L$_{1}$-norm-bi}  &
					\makecell{ASSL-bi}  &
					\makecell{SSL-bi (Ours)} 
					\\
					\includegraphics[width=\widthscalefive \textwidth]{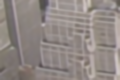}  &
					\includegraphics[width=\widthscalefive \textwidth]{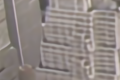}  &
					\includegraphics[width=\widthscalefive \textwidth]{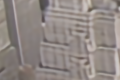}  &
					\includegraphics[width=\widthscalefive \textwidth]{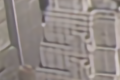}  &
					\includegraphics[width=\widthscalefive \textwidth]{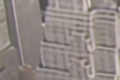}  &
					\includegraphics[width=\widthscalefive \textwidth]{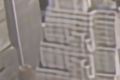} 
					\\ 
					EDVR-M  &
					\makecell{BasicVSR-uni}  &
					\makecell{BasicVSR-uni-lite}  &
					\makecell{L$_{1}$-norm-uni}  &
					\makecell{ASSL-uni}  &
					\makecell{SSL-uni (Ours)} 
				\end{tabular}
			\end{adjustbox}	
			\\
			\begin{adjustbox}{valign=t}
				\LARGE
				\begin{tabular}{c}
					\includegraphics[height=0.44\textwidth]{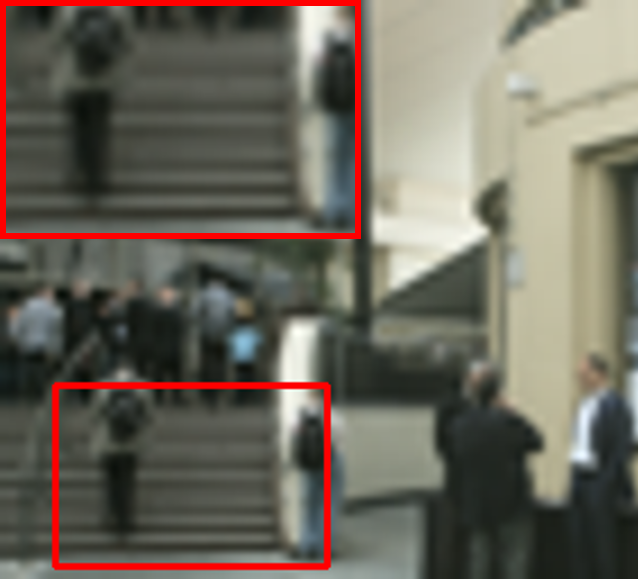} \\
					LR
				\end{tabular}
				
			\end{adjustbox}
			
			\begin{adjustbox}{valign=t}
				\LARGE
				\begin{tabular}{cccccc}
					\includegraphics[width=\widthscalefive \textwidth]{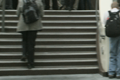}  &
					\includegraphics[width=\widthscalefive \textwidth]{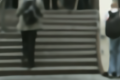}  &
					\includegraphics[width=\widthscalefive \textwidth]{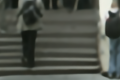}  &
					\includegraphics[width=\widthscalefive \textwidth]{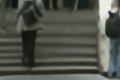}  &
					\includegraphics[width=\widthscalefive \textwidth]{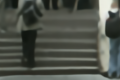}  &
					\includegraphics[width=\widthscalefive \textwidth]{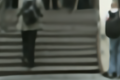} 
					\\
					HR  &
					\makecell{BasicVSR}  &
					\makecell{BasicVSR-lite}  &
					\makecell{L$_{1}$-norm-bi}  &
					\makecell{ASSL-bi}  &
					\makecell{SSL-bi (Ours)} 
					\\
					\includegraphics[width=\widthscalefive \textwidth]{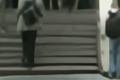}  &
					\includegraphics[width=\widthscalefive \textwidth]{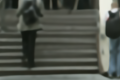}  &
					\includegraphics[width=\widthscalefive \textwidth]{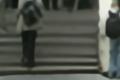}  &
					\includegraphics[width=\widthscalefive \textwidth]{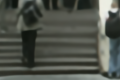}  &
					\includegraphics[width=\widthscalefive \textwidth]{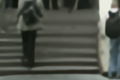}  &
					\includegraphics[width=\widthscalefive \textwidth]{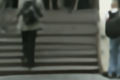} 
					\\ 
					EDVR-M  &
					\makecell{BasicVSR-uni}  &
					\makecell{BasicVSR-uni-lite}  &
					\makecell{L$_{1}$-norm-uni}  &
					\makecell{ASSL-uni}  &
					\makecell{SSL-uni (Ours)} 
				\end{tabular}
			\end{adjustbox}	
		\end{tabular}
	}
	\vspace{-3mm}
	\caption{Qualitative comparison between various VSR and pruning schemes on REDS4~\cite{REDS}, Vid4~\cite{Vid4}, and Vimeo90K-T~\cite{VSR6}, separately.}
	\label{fig:SR_show}
	\vspace{-4mm}
\end{figure*}

The qualitative results are shown in Fig.~\ref{fig:SR_show}. Our SSL-bi achieves the best visual quality containing more realistic details. These visual comparisons are consistent with the quantitative results, showing the superiority of SSL. SSL can learn to remove the redundant filters to compress a network to a smaller one while maintaining the most restoration ability. More visual results are given in supplementary.

\vspace{-2mm}
\section{Ablation Study}
\vspace{-1mm}
\noindent
\textbf{The Validation of Components in SSL.}
We conduct an ablation study to demonstrate the effectiveness of our SSL by progressively adding components. The results are shown in Tab.~\ref{tab:SSL}. SSL$_{1}$ uses the aligned pruning~\cite{ASSL} scheme for residual blocks, while SSL$_{4}$ adopts our RSC. Comparing SSL$_{1}$ and SSL$_{4}$, we can see that our RSC is superior to the advanced residual block pruning scheme. That is because RSC can break the pruning restrictions and preserve all information contained in feature channels for VSR. For SSL$_{2}$, we halve the number of filters in upsampling networks to keep the same model size as other models.  As we can see, SSL$_{3}$ outperforms SSL$_{2}$. This is because introducing the pruning scheme for the pixel-shuffle operation can increase the available pruning space in upsampling networks. Comparing SSL$_{4}$ and SSL$_{3}$, we can see that adopting TF can bring a 0.08 dB improvement, reducing the error accumulation of hidden states in the recurrent network after pruning. 

\begin{table}[t]
\LARGE
	\centering
	\caption{Validation of the components in our SSL. PSNR (dB) results evaluated on REDS4 \cite{REDS} ($4\times$). The backbone is BasicVSR \cite{BasicVSR}, and the pruning ratio is set to $0.5$. }
	\vspace{-3mm}
	\resizebox{0.9\linewidth}{!}{
		\begin{tabular}{l|c|ccc}
		    
			\toprule[0.2em]
			Methods & SSL$_{1}$ & SSL$_{2}$ & SSL$_{3}$  & SSL$_{4}$ (Ours) \\
			 \midrule
			Aligned Pruning \cite{ASSL} & \Checkmark     &       &       &         \\
			
			Residual Sparsity Connection &       & \Checkmark     & \Checkmark       & \Checkmark \\
			
			Pixel-Shuffle Pruning & \Checkmark     &       & \Checkmark     &     \Checkmark \\
			
			Temporal Finetuning & \Checkmark     &       &            & \Checkmark \\
			 \midrule
			PSNR (dB) & 30.86 & 30.82 & 30.98  & 31.06 \\
			\bottomrule[0.2em]
		\end{tabular}%
	}
	\label{tab:SSL}%
	\vspace{-6mm}
\end{table}%

\begin{figure}[tb]
	\large
	\centering
	\resizebox{0.9\linewidth}{!}{
	\includegraphics[width=1\linewidth]{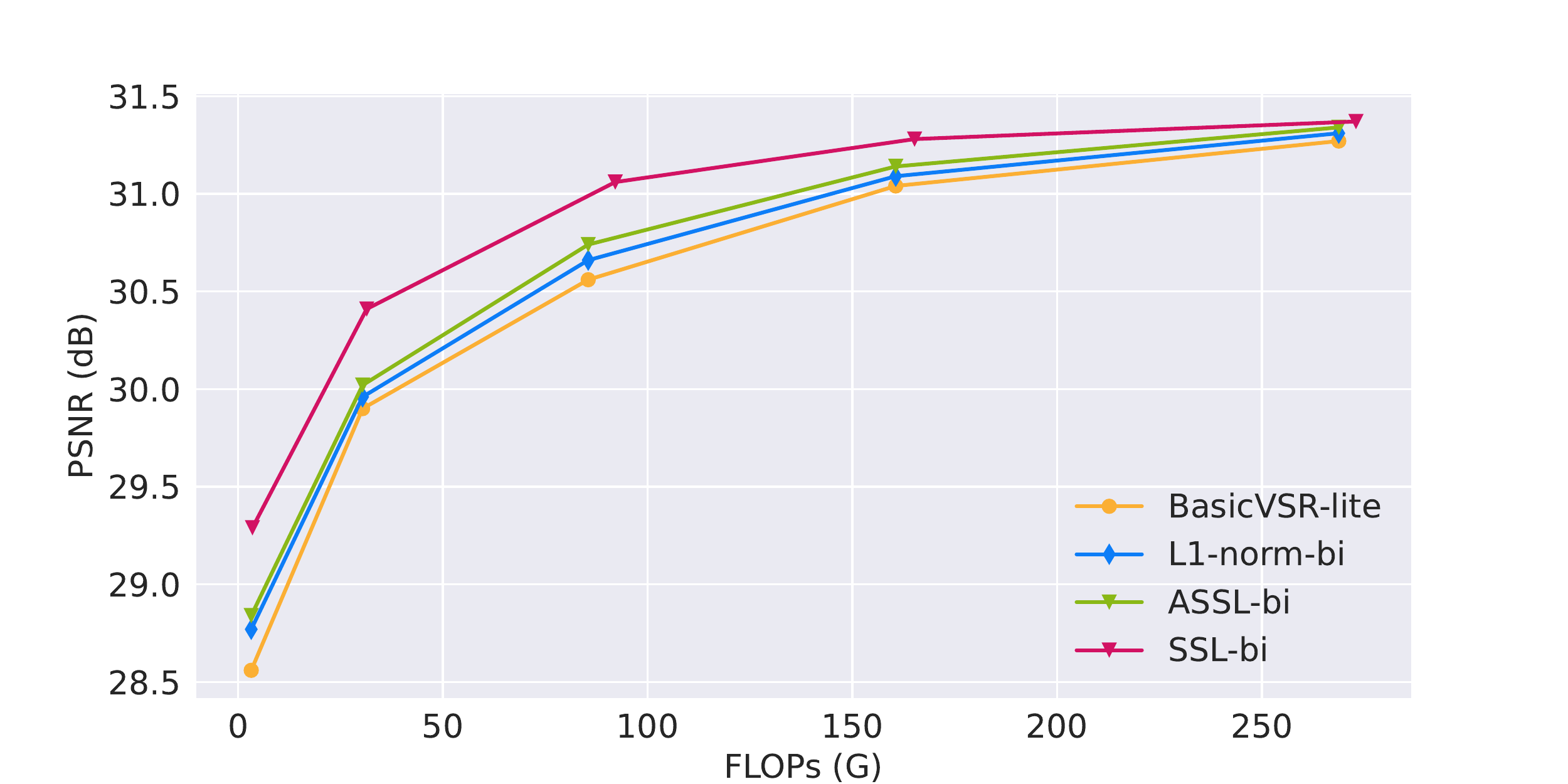}
	}
	\vspace{-4mm}
	\caption{PSNR (dB) comparison on REDS4 ($\times$4) between SSL and
	three other methods obtaining the \emph{same} small network. }
	\vspace{-7mm}
	\label{fig:pruning ratio}
\end{figure}

\noindent
\textbf{Pruning Methods with Various Pruning Ratios.}
To further demonstrate SSL's effectiveness, we compare it with widely used pruning schemes, including L$_{1}$-norm~\cite{PL1}, ASSL~\cite{ASSL} at different pruning ratios. We use these pruning schemes to obtain numerous submodels with different FLOPs. Besides, we also adjust the channels of BasicVSR to obtain BasicVSR-lite with different FLOPs. The results are shown in Fig.~\ref{fig:pruning ratio}. \textbf{(1)} Our SSL achieves the best performance compared with other pruning schemes at different pruning ratios and FLOPs. Note that SSL even surpasses the BasicVSR-lite in the same model size by 0.73 dB on submodels with around 3.2G FLOPs. This demonstrates the superiority of SSL on VSR. \textbf{(2)} With the pruning ratio increasing and FLOPs decreasing, the performance advantage brought by our SSL becomes more evident compared with BasicVSR-lite, L$_{1}$-norm and ASSL. 

\begin{figure*}[t]
	\newlength\fsdttwofigBD
	\setlength{\fsdttwofigBD}{-5mm}
	\scriptsize
	\centering

\resizebox{0.93\linewidth}{!}{
\begin{tabular}{ccc}
\hspace{-10mm} \includegraphics[height=0.23\linewidth]{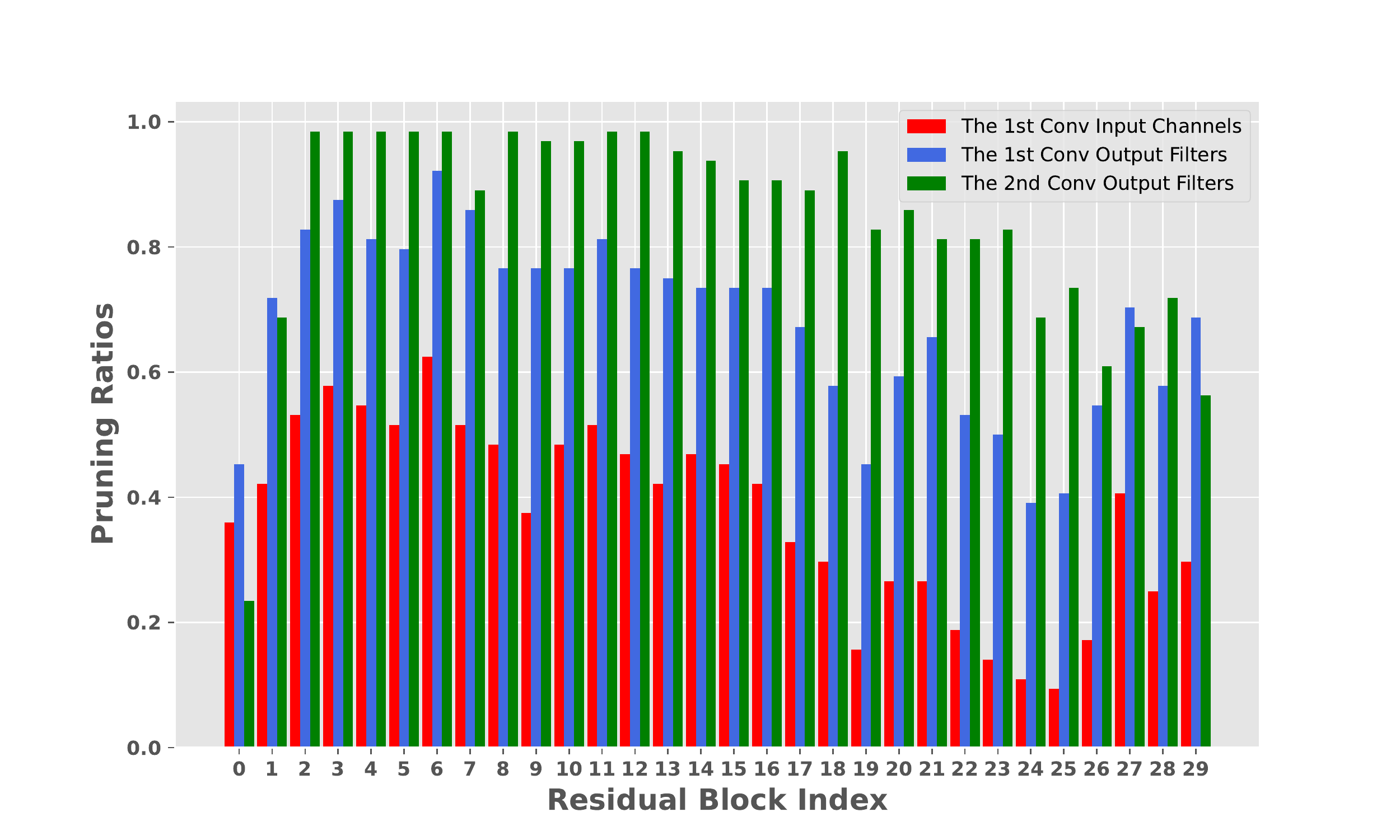}  & 
\hspace{-10mm} \includegraphics[height=0.23\linewidth]{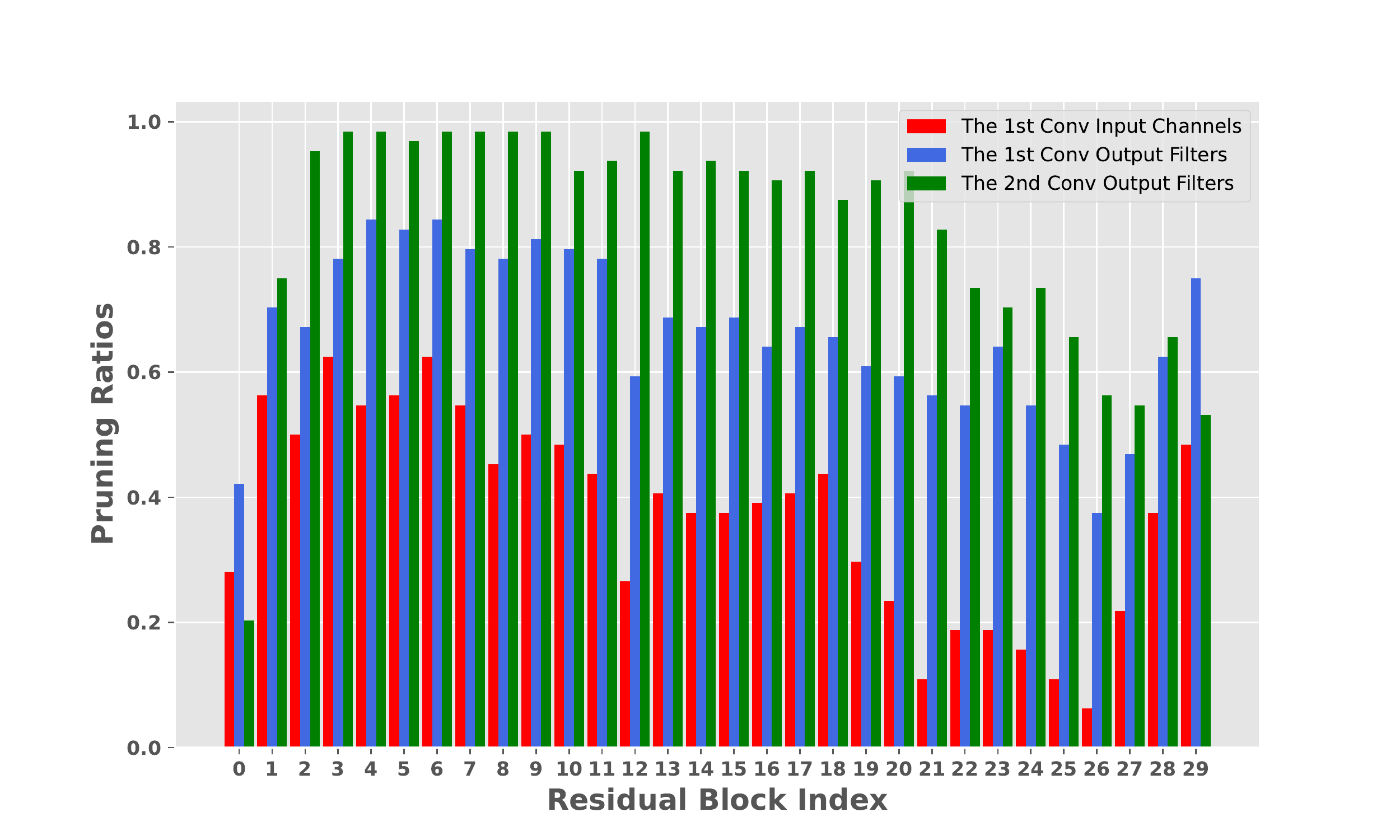}	 &
\hspace{-7mm} \includegraphics[height=0.2\linewidth]{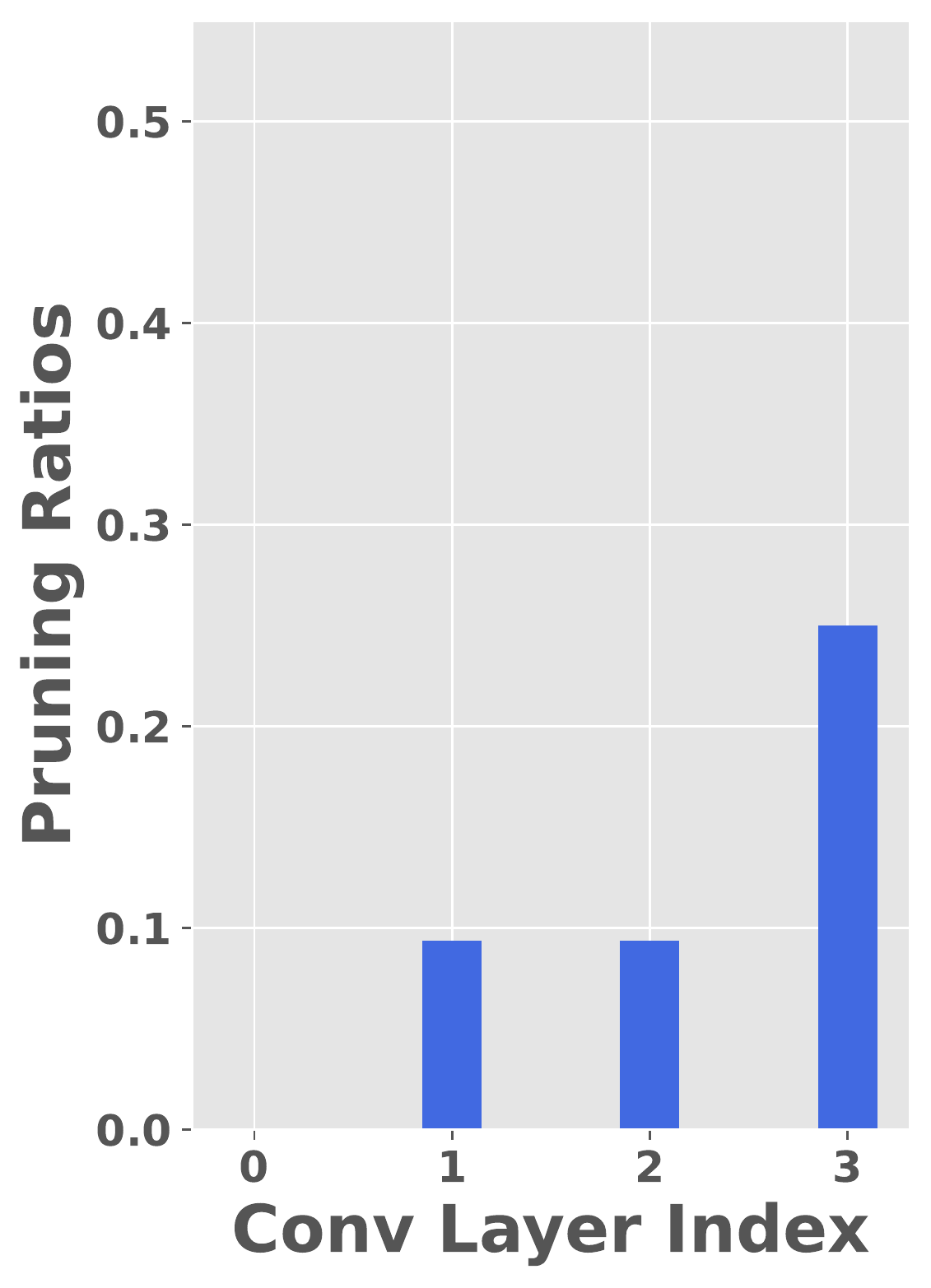}  \\
\hspace{-10mm} \shortstack{(a)  Forward  Network.}   &
\hspace{-10mm} \shortstack{(b) Backward  Network.}   &	
\hspace{-7mm} \shortstack{(c)  Upsampling Network.}   
\end{tabular}
}
\vspace{-3mm}
\caption{
 \textbf{(a)}, \textbf{(b)}, and \textbf{(c)} show the pruning ratios of Conv layers in forward, backward  and upsampling networks, respectively. }
 \vspace{-5mm}
\label{fig:hist}
\end{figure*}

\noindent
\textbf{Comparison with Different Pruning Criteria.}
We explore the influence of different pruning criteria on the pruned VSR model at different pruning ratios. Specifically, we select and remove the unimportant filters globally (namely, comparing all filters from all layers together) with minimum L$_{1}$-norm scores, which is expressed as ``Min + Global''. In addition, we select and remove the unimportant filters locally (namely, filters are compared with each other in the same layer, and each layer has the same pruning ratio) with maximum L$_{1}$-norm scores, which is expressed as ``Max + Local''. Similarly, we determine ``Max + Global'' and ``Min + local''. Furthermore, we randomly remove the unimportant filters as ``Rand''. Then, we compare all pruning criteria at $0.3$, $0.5$, and $0.7$ pruning ratios. The results are shown in Tab.~\ref{tab:criterion}. \textbf{(1)} The ``Min + Global'' pruning criterion achieves the best performance at different pruning ratios. It implies that the filters with minimum L$_{1}$-norm scores are relatively unimportant for VSR. Besides, this shows that the importance of filters in different layers are different and it is better to compare them together to select unimportant filters (pruning globally) for VSR. \textbf{(2)} The performance of ``Max + Global'' and ``Max + local'' are both inferior to ``Rand'' for the removal of more important filters. This implies that filters with large L$_{1}$-norm scores are more important than those with small ones for VSR networks.

\begin{table}[t]
\Large
	\centering
		\caption{PSNR (dB) comparison on REDS4 ($4\times$) for our pruning scheme (SSL) with different pruning criteria and pruning ratios. The unpruned model is BasicVSR~\cite{BasicVSR} baseline.}
		\vspace{-3mm}
		\resizebox{0.9\linewidth}{!}{
	\begin{tabular}{cccccc}
		\toprule[0.2em]
		\makecell{Pruning \\Ratios} & \makecell{Min  \\Global (Ours)} & \makecell{Max  \\Global} & \makecell{Min  \\Local} & \makecell{Max  \\Local} & Rand \\
		 \midrule
		0.3   & 31.28 & 30.59 & 31.22 & 30.97 & 31.10
 \\

		0.5   & 31.06 & 28.90 & 30.83 &30.38 &30.68
 \\

		0.7   & 30.41 & 25.89 & 30.20 & 29.51 & 30.09
 \\
		\bottomrule[0.2em]
	\end{tabular}%
	}
	\vspace{-7mm}
	\label{tab:criterion}%
\end{table}%

\noindent
\textbf{Pruning Ratios of Different Layers.}
We take BasicVSR pruned by SSL at $0.5$ pruning ratio as an example. We visualize pruning ratios in different layers and show results in Fig.~\ref{fig:hist}. \textbf{(1)} In the forward and backward networks, the pruning ratios of the first Conv input channels (corresponding to the $\boldsymbol{\gamma_{j-1}}$ in Fig.~\ref{fig:head} (d)) are lower than the pruning ratios of second Conv output filters (corresponding to the $\boldsymbol{\gamma_{j+1}}$ in Fig.~\ref{fig:head} (d)). It implies that 
VSR networks tend to aggregate information from the numerous input channels into several important output channels. \textbf{(2)} The residual blocks in deeper position of forward and backward networks tend to have minor pruning ratio. This means that residual blocks in deeper position contribute more to VSR. \textbf{(3)} The average pruning ratio of upsampling network is $0.2$ (less than $0.5$), indicating that upsampling network plays a quite important role in VSR performance. Previous works~\cite{EDSR,ESRGAN,RCAN,BasicVSR,EDVR, RLSP, PFNL} have paid much attention to the design of feature extraction modules. In latter VSR research, paying more attention to upsampling network design is likely to improve VSR performance more.

\begin{figure}[tb]
	\centering
	\resizebox{0.9\linewidth}{!}{
	\includegraphics[width=1\linewidth, height=0.55\linewidth]{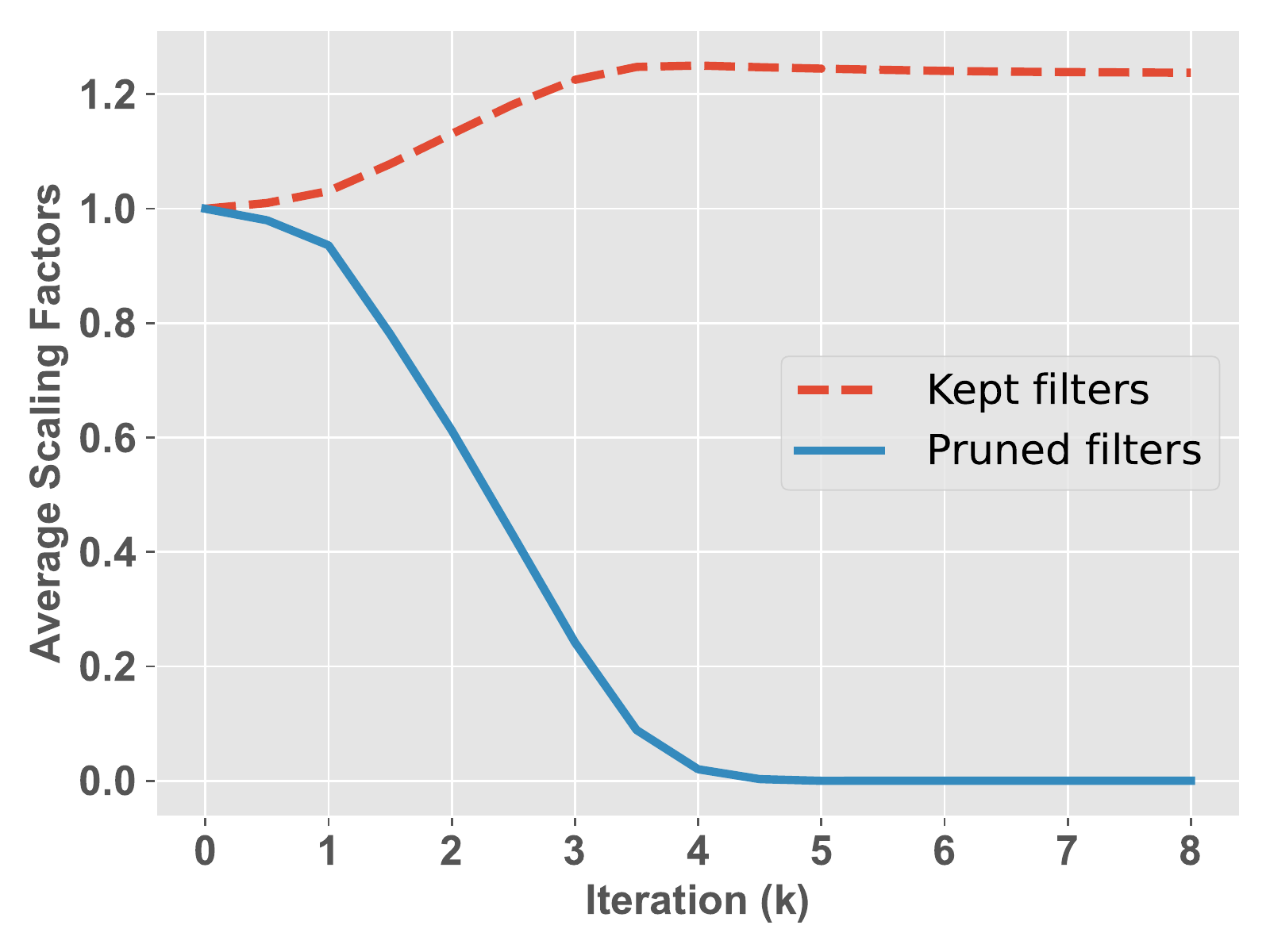} 
	}
	\vspace{-4mm}
	\caption{The SSL pruning process of Convs in the BasicVSR~\cite{BasicVSR}.}
	\vspace{-6mm}
	\label{fig:reg}
\end{figure}

\noindent
\textbf{Regularization Visualization.}
To understand how sparsity-inducing regularization works, we plot the average scaling factors in the BasicVSR~\cite{BasicVSR} during applying regularization at $0.5$ pruning ratio (Fig.~\ref{fig:reg}). The average scaling factor is split into two parts, pruned and kept. As seen, the average scaling factor $\gamma$ of the pruned filters decreases as the corresponding penalty term $\alpha_{\gamma}$ and iterations become larger. Besides, it is interesting that the average scaling factor of the kept filters will increase without any regularization term to enforce them to be larger. It means that, as the unimportant filters are removed, the network will strengthen the kept filters to compensate for the performance, which is similar to the compensation effect in the human brain.

\vspace{-2mm}
\section{Conclusion}
\vspace{-1mm}
In this work, we propose a structured pruning scheme called SSL for efficient VSR in resource-limited situations. Specifically, for the difficulty of pruning residual blocks of recurrent networks, we propose the RSC. Compared with previous pruning schemes for residual blocks, RSC does not have pruning restrictions as other pruning schemes and can fully utilize restoration information in all channels for better performance. In addition, for the pixel-shuffle operation in the upsampling network, we specially design a pruning scheme by grouping filters to guarantee the accuracy of channel-space conversion after pruning.  Furthermore, we propose Temporal Finetuning to reduce the error accumulation in recurrent networks. We apply SSL on the BasicVSR, and SSL achieves superior performance to that of recent SOTA methods, quantitatively and qualitatively.

{\small
\bibliographystyle{ieee_fullname}
\bibliography{egbib}
}

\end{document}